\documentclass[runningheads]{llncs}

\newif\ifanonymized
\anonymizedfalse   

\usepackage[T1]{fontenc}
\usepackage{marvosym}
\usepackage{graphicx,verbatim}
\usepackage{amsmath,amssymb}
\usepackage{booktabs}
\usepackage{bm}
\usepackage{float}
\usepackage{algorithm}
\usepackage{algpseudocode}
\usepackage[hidelinks]{hyperref}
\usepackage{color}

\begin{document}
	
	\title{ReCo-Diff: Residual-Conditioned Deterministic Sampling for Cold Diffusion in Sparse-View CT}
	\titlerunning{ReCo-Diff}
	\ifanonymized
	\author{Anonymous Authors}
	\authorrunning{Anonymous Author et al.}
	\institute{Anonymized Affiliation\\ \email{email@anonymized.com}}
	\else
	\author{
		Yong Eun Choi\inst{1} \and
		Hyoung Suk Park\inst{1} \and
		Kiwan Jeon\inst{1} \and
		Hyun-Cheol Park \inst{2} \and
		Sung Ho Kang\inst{1(\textrm{\Letter})}
	}
	\authorrunning{Y. E. Choi et al.}
	\institute{
		National Institute for Mathematical Sciences, Daejeon, 34047, Republic of Korea\\		
		\and
		Department of Computer Engineering, Korea National University of Transportation, Chungju, 27469, Republic of Korea\\
		\email{runits@nims.re.kr}
	}
	\fi
	
	\maketitle
	
	\begin{abstract}
		Cold and generalized diffusion models have recently shown strong potential for sparse-view CT reconstruction by explicitly modeling deterministic degradation processes. However, existing sampling strategies often rely on ad hoc sampling controls or fixed schedules, which remain sensitive to error accumulation and sampling instability. We propose \textbf{ReCo-Diff}, a residual-conditioned diffusion framework that leverages observation residuals through residual-conditioned self-guided sampling. At each sampling step, ReCo-Diff first produces a null (unconditioned) baseline reconstruction and then conditions subsequent predictions on the observation residual between the predicted image and the measured sparse-view input. This residual-driven guidance provides continuous, measurement-aware correction while preserving a deterministic sampling schedule, without requiring heuristic interventions. Experimental results demonstrate that ReCo-Diff consistently outperforms existing cold diffusion sampling baselines, achieving higher reconstruction accuracy, improved stability, and enhanced robustness under severe sparsity.		
		\ifanonymized
		Code will be released upon acceptance.
		\else
		The code is available at {\tt\small \url{https://github.com/choiyoungeunn/ReCo-Diff}}.
		\fi
	\end{abstract}

\section{Introduction}

Sparse-view computed tomography (CT) aims to reconstruct diagnostically reliable images from angularly under-sampled projection data in order to reduce radiation dose and acquisition time~\cite{brenner2007computed,miller1983alara}.
However, angular subsampling induces structured streak artifacts that fundamentally differ from additive Gaussian noise~\cite{sidky2008image}, rendering the reconstruction problem highly ill-posed and challenging for conventional imaging models.

Classical reconstruction methods, such as filtered backprojection (FBP) and iterative techniques based on total variation or compressed sensing, provide theoretical guarantees but often suffer from severe streak artifacts or over-smoothing under extreme sparsity.
Learning-based approaches have substantially improved sparse-view CT reconstruction by leveraging data-driven priors, operating in the image domain, sinogram domain, or both~\cite{jin2017deep,lee2018deep,lin2019dudonet,wu2021drone}.
Despite their success, most feed-forward networks are trained for fixed sparsity levels and lack robustness to varying view configurations or iterative error accumulation, often producing over-smoothed reconstructions under severe undersampling~\cite{li2023learning,ma2023freeseed}.

Diffusion models have recently emerged as powerful generative priors for inverse imaging problems, including sparse-view CT reconstruction~\cite{chung2022diffusion,chung2023direct,chung2022improving,he2024solving,song2021scorebased}.
While standard denoising diffusion probabilistic models rely on stochastic Gaussian noise corruption, generalized and cold diffusion frameworks extend diffusion models to arbitrary and deterministic degradation processes, making them better suited for modeling structured artifacts arising from angular subsampling~\cite{bansal2023cold}.
Building on this formulation, the Cross-view Generalized Diffusion Model (CvG-Diff) explicitly models sparse-view CT reconstruction as a generalized diffusion process with a physics-based degradation operator, and introduces Error-Propagating Composite Training (EPCT) and Semantic-Prioritized Dual-Phase Sampling to mitigate artifact propagation during multi-step sampling~\cite{chen2024cvgdiff}.

Despite these advances, sampling stability remains a critical challenge in cold and generalized diffusion models for sparse-view CT.
Iterative restoration under deterministic degradation inherently accumulates reconstruction errors, which are repeatedly injected into subsequent states.
To address this issue, existing approaches typically employ fixed sampling schedules or heuristic control mechanisms, such as structural similarity index measure (SSIM)-based adaptive resets, to suppress error propagation~\cite{chen2024cvgdiff}.
However, these strategies suffer from several limitations: (i) reset decisions are not directly tied to measurement consistency, (ii) repeated resets introduce additional computational overhead and non-deterministic sampling paths, and (iii) sampling stability remains sensitive to hyperparameter choices. To address these issues, recent studies have incorporated conditioning mechanisms into diffusion-based models to enhance controllability in inverse problems, including posterior sampling and Schrödinger Bridge formulations~\cite{chung2022diffusion,chung2023direct,liu20232}. These approaches indicate that structured conditioning can systematically guide sampling trajectories beyond fixed schedules or heuristic control strategies.

In this work, we propose a residual-conditioned self-guided sampling strategy for cold diffusion in sparse-view CT.
The proposed method explicitly conditions sampling on observation residuals,
enabling continuous, physics-aware error correction under a fixed and deterministic sampling schedule.
It is conceptually related to classifier-free guidance (CFG)~\cite{ho2022cfg} in its use of an unconditional baseline
followed by corrective guidance, while remaining fully deterministic without conditional–unconditional branching.

\begin{figure}[t]
	\centering
	\includegraphics[width=1.0\linewidth]{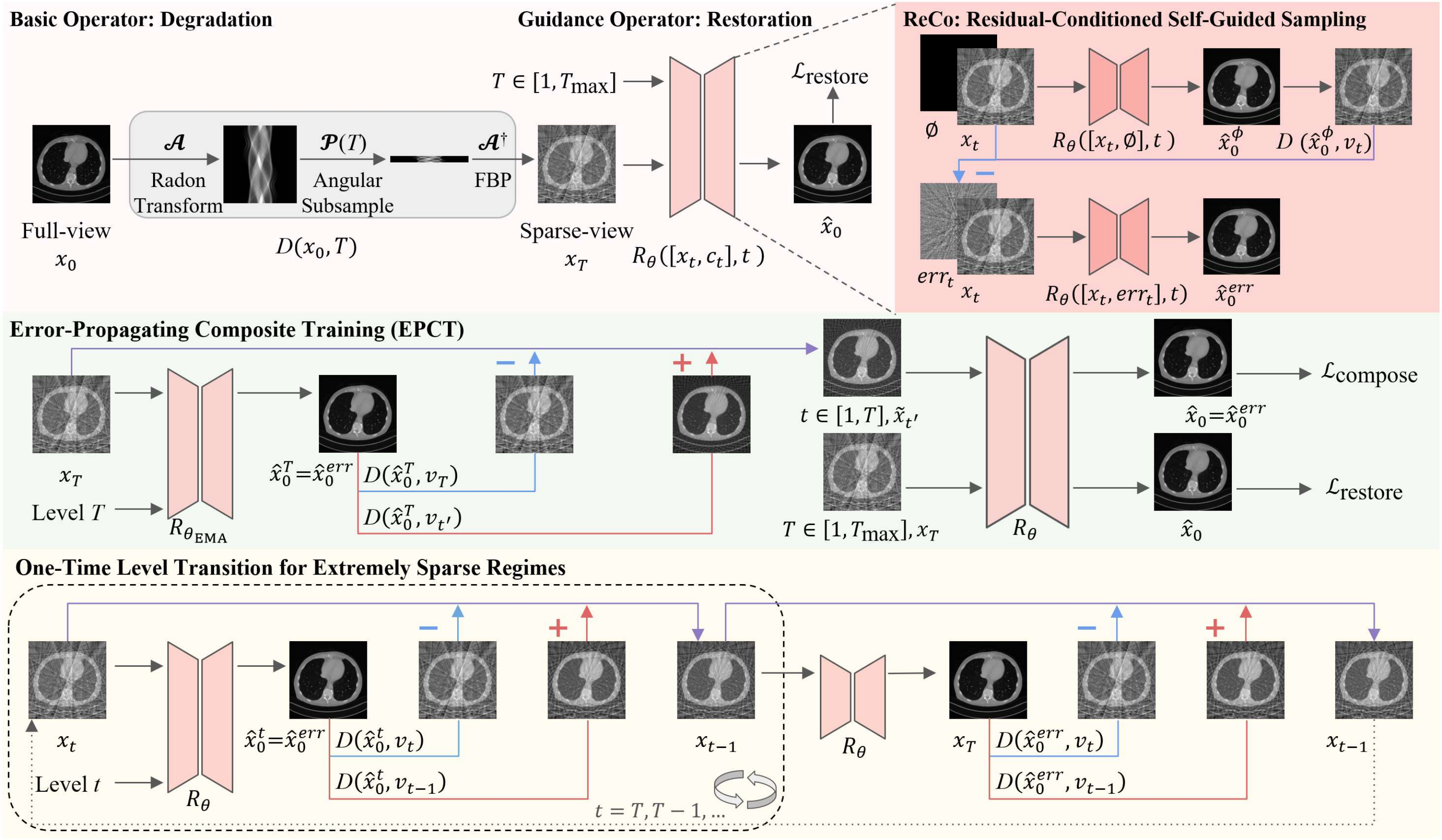}
	\caption{Overview of the proposed ReCo-Diff.}
	\label{fig:overview}
\end{figure}

\section{Methodology}

\subsection{Background: Cold Diffusion for Sparse-View CT}

We consider a generalized diffusion framework~\cite{bansal2023cold} in which a clean image $x_0$
is deterministically degraded to a severity level $t$ via an operator $D$:
\begin{equation}
	x_t = D(x_0,t), \qquad D(x_0,0)=x_0.
\end{equation}
A restoration network $R_\theta$ is trained to invert this degradation by minimizing
\begin{equation}
	\mathcal{L}_{\text{restore}} = \|R_\theta(x_t,t) - x_0\|_2,\quad x_t=D(x_0,t).
	\label{eq:loss_restore}
\end{equation}

During inference, restoration and degradation are alternated as
\begin{align}
	\hat{x}_{0,t} &= R_\theta(x_t,t), \\
	x_{t-1} &= x_t - D(\hat{x}_{0,t},t) + D(\hat{x}_{0,t},t-1),
\end{align}
iterated from $t=T$ to $t=1$.

For sparse-view CT, the degradation operator is defined as
\begin{equation}
	x_v = D(x_0,v) = \mathcal{A}^\dagger \mathcal{P}(v)\mathcal{A}x_0,
\end{equation}
where $\mathcal{A}$ denotes the Radon transform,
$\mathcal{P}(v)$ applies angular subsampling,
and $\mathcal{A}^\dagger$ is filtered backprojection~\cite{chen2024cvgdiff}.
We use a discrete timestep index $t$ to denote the sampling step, and associate each step $t$ with a corresponding view level $v_t$ under a unified degradation schedule. Accordingly, the degraded input at step $t$ can be written as $x_{v_t} = D(x_0, v_t)$.

\subsection{Training with EPCT and Residual Conditioning}

We adopt the Error-Propagating Composite Training (EPCT) strategy from CvG-Diff~\cite{chen2024cvgdiff}
to explicitly train the restoration network under error-propagated intermediate states.
Our training objective consists of two terms:
(i) a standard restoration loss $\mathcal{L}_{\text{restore}}$ and
(ii) a composite loss $\mathcal{L}_{\text{compose}}$ that simulates multi-step artifact propagation.

\paragraph{Concat-form conditioning.}
Residual information is incorporated by concatenating the conditioning signal to the degraded input state along the channel dimension.
Specifically, the restoration network takes the concatenated input $[x_t, c_t]$ at level $t$, where $x_t$ denotes the degraded CT image at level $t$,
$c_t=\emptyset$ corresponds to the null (unconditioned) baseline,
and $c_t=err_t$ provides residual-based guidance.

\paragraph{Direct restoration loss.}
Given a clean CT slice $x_0$ and a sampled view level $v_t$, we generate the sparse-view input
\begin{equation}
	x_t = D(x_0, v_t).
\end{equation}
We first compute a null (unconditioned) baseline prediction
\begin{equation}
	\hat{x}_{0,t}^{\phi} = R_\theta([x_t,\emptyset], t),
\end{equation}
and define the observation residual as
\begin{equation}
	err_t = \mathcal{N}\!\left(x_t - D(\hat{x}_{0,t}^{\phi}, v_t)\right),
	\label{eq:train_residual}
\end{equation}

where $\mathcal{N}(\cdot)$ denotes a bounded residual normalization implemented using a hyperbolic tangent,
which maps the raw observation mismatch to a fixed range.
This normalization stabilizes residual-conditioned training and sampling
by preventing residual explosion at early timesteps and avoiding channel dominance in concatenated conditioning.
It also helps reduce training–inference mismatch by keeping the scale of residual guidance consistent across sampling steps.
The residual-conditioned prediction is then obtained by

\begin{equation}
	\hat{x}_{0,t}^{err} = R_\theta([x_t, err_t], t).
\end{equation}
The direct restoration loss is defined as
\begin{equation}
	\mathcal{L}_{\text{restore}} = \|\hat{x}_{0,t}^{err} - x_0\|_2.
\end{equation}

\paragraph{Composite loss for error propagation (EPCT).}
To expose the model to propagated artifacts, we additionally construct an error-propagated intermediate state
using an EMA teacher network.
Specifically, we sample a target level $v_T$ and an intermediate level $v_{t'} < v_T$.
Let $x_T = D(x_0, v_T)$ and obtain the teacher prediction
\begin{equation}
	\hat{x}_{0,T} = R_{\theta_{\mathrm{EMA}}}([x_T, \emptyset], T).
\end{equation}
We then synthesize an error-propagated intermediate state
\begin{equation}
	\tilde{x}_{t'} = x_T - D(\hat{x}_{0,T}, v_T) + D(\hat{x}_{0,T}, v_{t'}).
\end{equation}
Residual conditioning is applied in the same concat form:
\begin{align}
	\hat{x}_{0,t'}^{\phi} &= R_\theta([\tilde{x}_{t'},\emptyset], t'),\\
	\tilde{err}_{t'} &= \mathcal{N}\!\left(\tilde{x}_{t'} - D(\hat{x}_{0,t'}^{\phi}, v_{t'})\right),\\
	\hat{x}_{0,t'}^{err} &= R_\theta([\tilde{x}_{t'},\tilde{err}_{t'}], t').
\end{align}
The composite loss is defined as
\begin{equation}
	\mathcal{L}_{\text{compose}} = \|\hat{x}_{0,t'}^{err} - x_0\|_2.
\end{equation}

\paragraph{Overall objective and optimization.}
We optimize
\begin{equation}
	\mathcal{L}_{\text{train}} = \mathcal{L}_{\text{restore}} + \mathcal{L}_{\text{compose}}.
\end{equation}
In practice, we perform two sequential gradient updates within each iteration:
first using $\mathcal{L}_{\text{restore}}$ on directly generated sparse-view inputs,
and then using $\mathcal{L}_{\text{compose}}$ on EPCT-synthesized error-propagated states.
This two-stage update improves robustness to artifact accumulation and aligns training with multi-step inference.

\begin{figure}[t]
	\centering
	\begin{minipage}[t]{0.48\columnwidth}
		\begin{algorithm}[H]
			\caption{Residual-Conditioned Training (Direct Restoration)}
			\label{alg:training}
			\begin{algorithmic}[1]
				\Require Clean CT slices $\{x_0\}$, degradation operator $D$, restoration network $R_\theta$
				\For{each iteration}
				\State Sample $t, v_t$
				\State $x_t \leftarrow D(x_0, v_t)$
				\State $\hat{x}_{0,t}^{\phi} \leftarrow R_\theta([x_t,\emptyset], t)$
				\State $err_t \leftarrow \mathcal{N}(x_t - D(\hat{x}_{0,t}^{\phi}, v_t))$
				\State $\hat{x}_{0,t}^{err} \leftarrow R_\theta([x_t, err_t], t)$
				\State Update $\theta$ using $\mathcal{L}_{\text{restore}}$
				\EndFor
			\end{algorithmic}
		\end{algorithm}
	\end{minipage}
	\hfill
	\begin{minipage}[t]{0.48\columnwidth}
		\begin{algorithm}[H]
			\caption{Residual-Conditioned Self-Guided Sampling}
			\label{alg:residual_guided_sampling}
			\begin{algorithmic}[1]
				\Require $x_T$, $\{v_T,\ldots,v_0\}$, $R_\theta$, $D$
				\State $x \leftarrow x_T$
				\For{$t=T,\ldots,1$}
				\State $\hat{x}_{0,t}^{\phi} \leftarrow R_\theta([x,\emptyset], t)$
				\State $err_t \leftarrow \mathcal{N}(x - D(\hat{x}_{0,t}^{\phi}, v_t))$
				\State $\hat{x}_{0,t}^{err} \leftarrow R_\theta([x, err_t], t)$
				\State $x \leftarrow x - D(\hat{x}_{0,t}^{err}, v_t) + D(\hat{x}_{0,t}^{err}, v_{t-1})$
				\EndFor
				\State \Return $\hat{x}_0 \leftarrow x$
			\end{algorithmic}
		\end{algorithm}
	\end{minipage}
\end{figure}

\begin{figure}[t]
	\centering
	\includegraphics[width=0.9\linewidth]{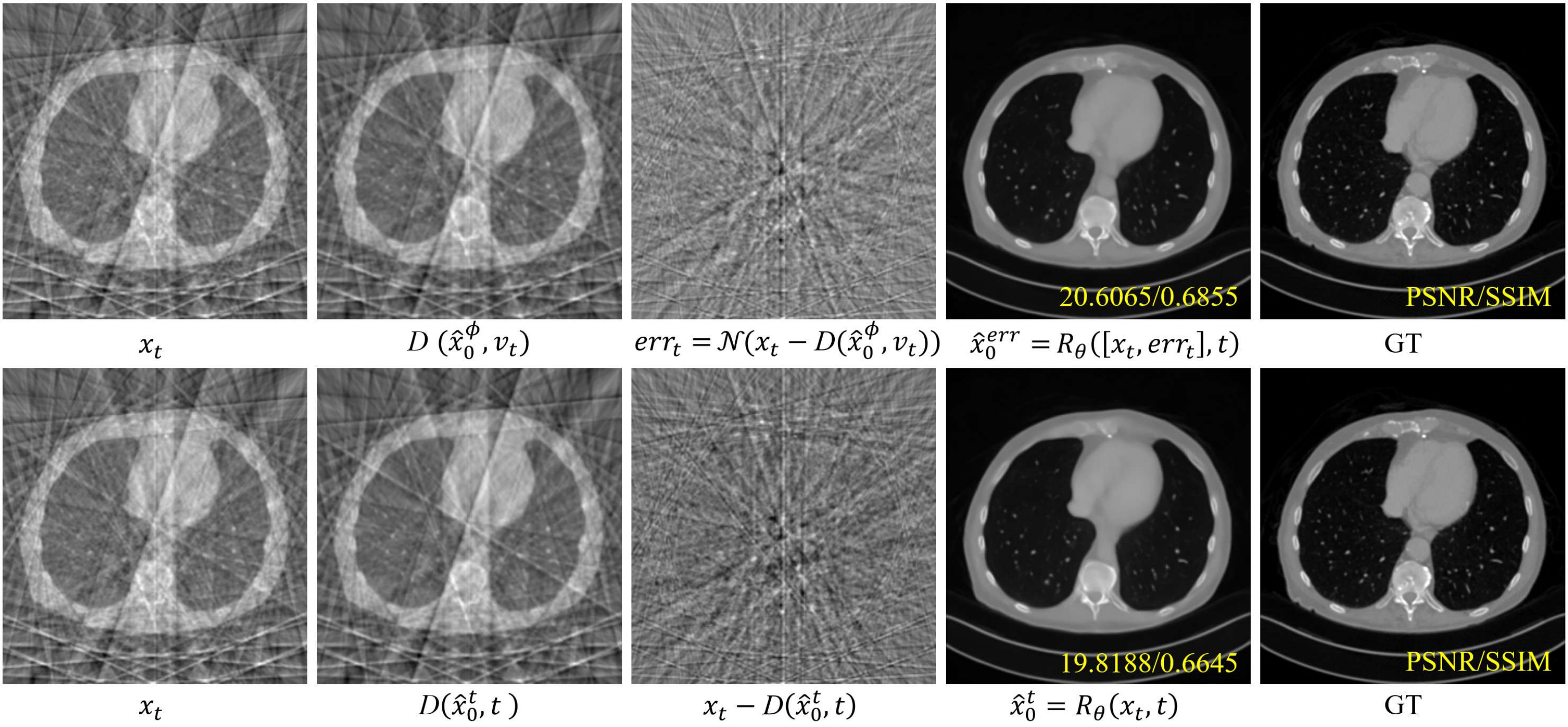}
	\caption{Reconstruction differences across error accumulation strategies. 
		\textbf{Top:} Proposed Residual-Conditioned Self-Guided Sampling. 
		\textbf{Bottom:} Sampling process of conventional cold diffusion.}
	\label{fig:proposed_reco}
\end{figure}

\subsection{Residual-Conditioned Self-Guided Sampling}

During inference, we employ a residual-conditioned self-guided sampling strategy
that replaces heuristic reset mechanisms with observation-aware guidance.
At each sampling step, a null (unconditioned)  baseline is computed to estimate an observation residual,
which is then used as a concatenated conditioning signal.

Given the current degraded state $x_t$, we compute
\begin{align}
	\hat{x}_{0,t}^{\phi} &= R_\theta([x_t,\emptyset], t),\\
	err_t &= \mathcal{N}\!\left(x_t - D(\hat{x}_{0,t}^{\phi}, v_t)\right),\\
	\hat{x}_{0,t}^{err} &= R_\theta([x_t, err_t], t),
\end{align}
and update the degraded state as
\begin{equation}
	x_{t-1} = x_t - D(\hat{x}_{0,t}^{err}, v_t) + D(\hat{x}_{0,t}^{err}, v_{t-1}),
\end{equation}
which continuously suppresses error accumulation while preserving a deterministic sampling schedule.

\paragraph{One-Time Level Transition for Extremely Sparse Regimes.}
In extremely sparse regimes (e.g., 18-view),
observation residuals at early sampling steps may be unreliable.
To stabilize initialization, we optionally perform a one-time level transition
after a small number of warm-up steps.
Specifically, we re-parameterize the degraded state as
\begin{equation}
	x_{t-1} \leftarrow x_T - D(\hat{x}_{0,t}^{err}, v_t) + D(\hat{x}_{0,t}^{err}, v_{t-1}),
\end{equation}
and continue sampling from level $v_{t-1}$ using the same residual-guided update rule.
This transition is performed only once and does not rely on heuristic criteria.

\paragraph{CFG-inspired interpretation.}
The proposed residual-conditioned sampling can be conceptually related to guidance mechanisms in diffusion models.
Unlike CFG~\cite{ho2022cfg}, which combines conditional and unconditional score predictions under Gaussian noise assumptions, our method derives guidance directly from observation residuals in image space.
These residuals explicitly reflect measurement inconsistency between the current reconstruction and the sparse-view input, enabling physically grounded correction without heuristic scaling or conditional--unconditional branching.

\begin{figure}[t]
	\centering
	\includegraphics[width=1.0\linewidth]{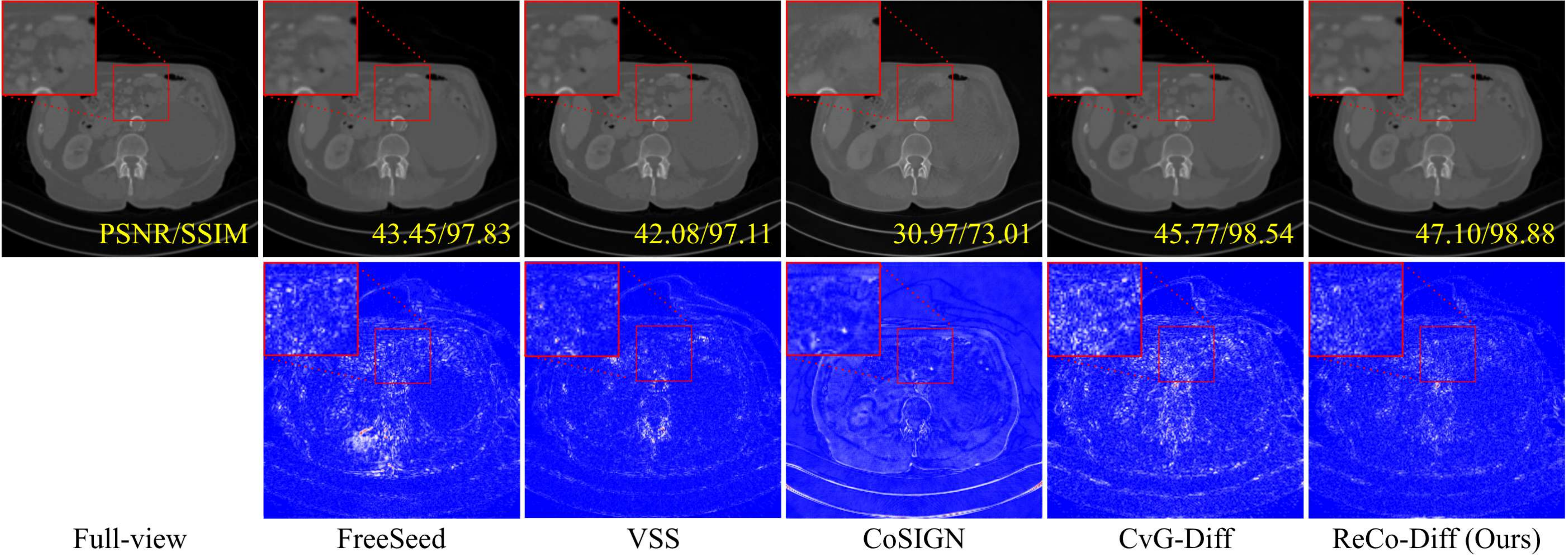}
	\caption{Visual comparison of different methods. Reconstructions are obtained with 72 target views and displayed using a window of $[-1000,\,2000]$ Hounsfield Unit (HU). Red regions denote higher errors.}
	\label{fig:quality_compare}
\end{figure}

\begin{figure}[t]
	\centering
	\includegraphics[width=0.9\linewidth]{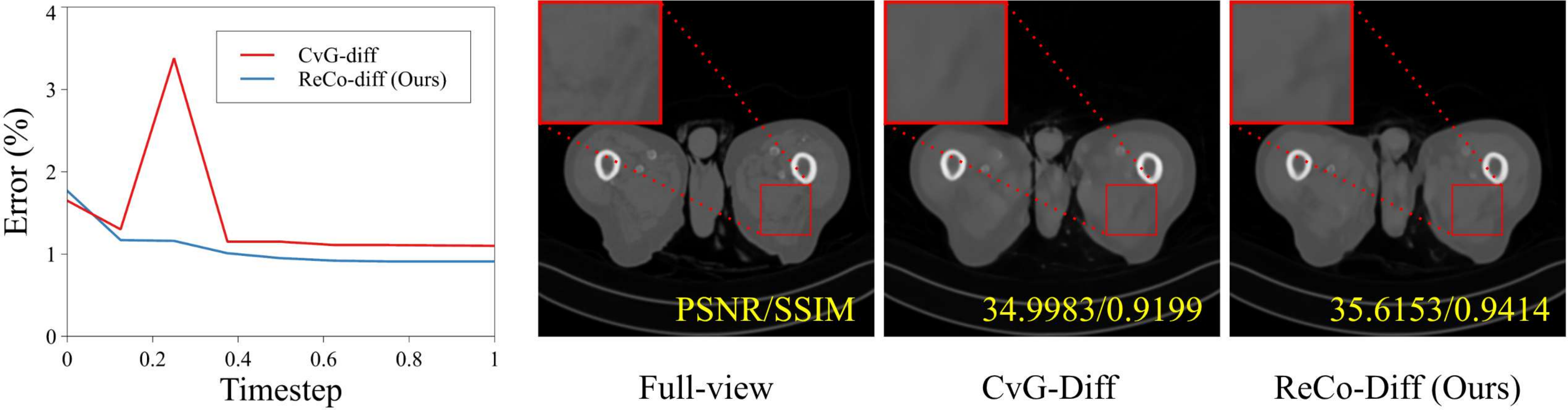}
	\caption{Comparison with CvG-Diff under different SSIM-based reset conditions.
		CvG-Diff triggers adaptive resetting when the SSIM value falls below a predefined threshold during sampling.
		In contrast, the proposed method demonstrates a consistently decreasing trend of observation residuals across sampling steps and achieves superior reconstruction quality in practice.}
	\label{fig:err_compare}
\end{figure}

\begin{table}[t]
	\centering
	\caption{Comparison results on the AAPM-LDCT dataset across different view settings.
		Mean results of RMSE [HU], PSNR [dB], SSIM [\%], and average reconstruction time at the 18-view setting [s] are reported.
		NFE denotes the number of network function evaluations in multi-step reconstruction.
		$\dagger$ indicates an ablation setting where the proposed method is evaluated without the one-time level transition strategy.
		Lower RMSE and higher PSNR/SSIM indicate better performance.
		The best result is shown in bold.}
	\label{tab:quant_views}
	\small
	\resizebox{\textwidth}{!}{
		\begin{tabular}{lccc ccc ccc c}
			\toprule
			Method (NFE) &
			\multicolumn{3}{c}{18-view} &
			\multicolumn{3}{c}{36-view} &
			\multicolumn{3}{c}{72-view} &
			Time \\ 
			\cmidrule(lr){2-4}\cmidrule(lr){5-7}\cmidrule(lr){8-10}
			& RMSE & PSNR & SSIM
			& RMSE & PSNR & SSIM
			& RMSE & PSNR & SSIM
			& \\
			\midrule
			FreeSeed~\cite{ma2023freeseed} (1) &
			50.69 & 35.49 & 95.21 &
			23.17 & 42.28 & 96.87 &
			18.95 & 44.03 & 98.01 &
			\textbf{0.09} \\
			VSS~\cite{he2024solving} (1000) &
			52.73 &	35.17 &	90.98 &
			32.87 &	39.34 &	95.15 &
			24.19 &	41.95 &	97.12 &
			264.71 \\
			CoSIGN~\cite{cosign} (10) &
			76.95 & 31.84 & 86.31 &
			53.73 & 34.96 & 89.67 & 
			38.37 & 37.87 & 93.20 &
			1.82 \\
			CvG-Diff~\cite{chen2024cvgdiff} (10) &
			36.65 & 38.33 & 95.18 &
			24.67 & 41.77 & 97.05 &
			15.77 & 45.63 & 98.54 &
			0.69 \\
			ReCo-Diff (16)~$\dagger$ &
			37.14 & 38.21 & 95.20 &
			23.35 & 42.22 & 97.33 &
			14.10 & 46.60 & 98.81 &
			0.76 \\
			\textbf{ReCo-Diff} (18) &
			\textbf{35.75} & \textbf{38.54} & \textbf{95.42} &
			\textbf{22.45} & \textbf{42.57} & \textbf{97.45} &
			\textbf{13.50} & \textbf{46.98} & \textbf{98.89} &
			0.86 \\
			\bottomrule
		\end{tabular}%
	}
\end{table}

\section{Experiments}

We evaluate the proposed method on the AAPM Low-Dose CT dataset~\cite{aapm16},
which contains 5,936 CT slices from 10 patients.
Following standard protocols, the dataset is divided into 5,410 training slices and 526 test slices.
Sparse-view projections are simulated using a fan-beam geometry implemented with the TorchRadon toolbox~\cite{ronchetti2020torchradon}.
Reconstruction experiments are conducted under three target view settings: 18, 36, and 72 views.
The restoration network is a diffusion U-Net with residual blocks and a base channel width of 128.
A unified degradation schedule is adopted to support reconstruction across all sparsity levels using a single model.
Training is performed using the Adam optimizer~\cite{diederik2014adam} with exponential moving average (EMA) stabilization.
Reconstruction quality is evaluated using root mean squared error (RMSE), peak signal-to-noise ratio (PSNR), and SSIM.

\subsection{Results}

We compare our method with FreeSeed~\cite{ma2023freeseed},
VSS~\cite{he2024solving}, CoSIGN~\cite{cosign},
and CvG-Diff~\cite{chen2024cvgdiff}.
We evaluate the proposed residual-conditioned self-guided sampling
for sparse-view CT reconstruction under different target view settings.
Quantitative results in Table~\ref{tab:quant_views} show that our method
consistently outperforms existing approaches across all view configurations,
with larger gains under severe sparsity.
Qualitative comparisons in Fig.~\ref{fig:quality_compare} demonstrate
reduced streak artifacts and better preservation of anatomical structures.
Furthermore, the error maps and timestep-wise curves in
Fig.~\ref{fig:err_compare} indicate more stable error trajectories
than SSIM-based reset strategies, improving stability
in multi-step reconstruction.

\section{Discussion}

ReCo-Diff provides a principled alternative to heuristic sampling control in cold diffusion by leveraging residual-conditioned guidance derived from observation residuals at each sampling step.
The model learns to correct errors in a continuous and physics-aware manner by directly conditioning on observation residuals.
This approach improves stability, reduces hyperparameter sensitivity, and avoids computational overhead associated with reset mechanisms.
Unlike CFG in noise-based diffusion, the proposed residual guidance operates in a deterministic image space, where the magnitude of the residual naturally reflects the severity of measurement inconsistency, eliminating the need for an explicit guidance scale.

\section{Conclusion}

We introduced a residual-conditioned self-guided sampling strategy for cold diffusion in sparse-view CT.
Our method replaces heuristic reset rules with observation-aware guidance, enabling stable and robust reconstruction under deterministic degradation.
The proposed approach generalizes CFG principles to cold diffusion and may serve as a practical alternative for stable sampling in deterministic inverse problems.

\ifanonymized

\else
\subsubsection{\ackname}
This work was supported in part by the National Research Foundation of Korea (NRF) Grant funded by Korean Government [Ministry of Science and ICT (MSIT)] under Grant RS-2024-00338504, and in part by the National Institute for Mathematical Sciences (NIMS) funded by Korean Government under Grant NIMS-B25910000.

\subsubsection{Disclosure of Interests.}
The authors have no competing interests to declare that are relevant to the content of this article.
\fi

\bibliographystyle{splncs04}
\bibliography{ref}

\end{document}